\documentclass[runningheads]{llncs}
\usepackage{graphicx}
\usepackage{paralist}
\usepackage{CJKutf8}
\usepackage{multirow}
\usepackage{subfigure}

\usepackage{array,tabularx}
\usepackage{color} 
\definecolor{category}{RGB}{112,48,160}
\definecolor{aspect}{RGB}{68,114,196}
\definecolor{opinion}{RGB}{159,34,0}
\definecolor{sentiment}{RGB}{255,34,0}
\newcommand{\blue}[1]{{\color{blue}{#1}}}

\usepackage{amsmath,amsfonts,bm}
\usepackage{mathtools}

\def\eqref#1{equation~\ref{#1}}

\def\1{\bm{1}}

\DeclareMathAlphabet{\mathsfit}{\encodingdefault}{\sfdefault}{m}{sl}
\SetMathAlphabet{\mathsfit}{bold}{\encodingdefault}{\sfdefault}{bx}{n}

\begin{document}
\title{An Empirical Study of Benchmarking Chinese Aspect Sentiment Quad Prediction}
%
%
\author{Junxian Zhou\inst{1} \and
Haiqin Yang\inst{2}\orcidID{0000-0001-5453-476X}\thanks{Corresponding author} \and
Junpeng Ye\inst{2} \and 
Yuxuan He\inst{1} \and 
Hao Mou\inst{1} 
}
\authorrunning{J. Zhou et al.}
%
\institute{DataStory, Guangzhou, China \\\email{\texttt{\{junius, heyuxuan, mouhao, junbo\}@datastory.com.cn}} \and
International Digital Economy Academy, Shenzhen, China\\
hqyang@ieee.org, yejunpeng@idea.edu.cn}
\maketitle              
\begin{abstract}


Aspect sentiment quad prediction (ASQP) is a critical subtask of aspect-level sentiment analysis.  Current ASQP datasets are characterized by their small size and low quadruple density, which hinders technical development.  To expand capacity, we construct two large Chinese ASQP datasets crawled from multiple online platforms. The datasets hold several significant characteristics: larger size (each with 10,000+ samples) and rich aspect categories, more words per sentence, and higher density than existing ASQP datasets.  Moreover, we are the first to evaluate the performance of Generative Pre-trained Transformer (GPT) series models on ASQP and exhibit potential issues. The experiments with state-of-the-art ASQP baselines underscore the need to explore additional techniques to address ASQP, as well as the importance of further investigation into methods to improve the performance of GPTs.  



\end{abstract}
\section{Introduction}\label{sec:intro}

Aspect-based sentiment analysis (ABSA), as a fine-grained opinion mining or sentiment analysis task, is significant because it can provide valuable insights for businesses and organizations to improve their products or services~\cite{DBLP:series/synthesis/2012Liu,DBLP:conf/semeval/PontikiGPPAM14,DBLP:journals/corr/abs-2203-01054}.  A complete ABSA structure consists of four main elements: (1) {\em aspect category} ($c$): the type of the concerned aspect; (2) {\em aspect term} ($a$): the opinion target which is explicitly or implicitly mentioned in the given text; (3) {\em opinion term} ($o$): the sentiment towards the aspect; and (4) {\em sentiment polarity} ($s$): the sentiment orientation.  Various methods, such as aspect term extraction (ATE)~\cite{DBLP:conf/acl/HeLND17,DBLP:conf/emnlp/LiL17}, aspect category detection (ACD)~\cite{DBLP:conf/acl/HeLND17,DBLP:conf/emnlp/0030TCL021}, aspect-opinion pairwise extraction (AOPE)~\cite{DBLP:conf/emnlp/WuWP20,DBLP:journals/taslp/YuJX19}, aspect-category sentiment analysis (ACSA)~\cite{DBLP:conf/coling/CaiTZYX20,DBLP:conf/emnlp/DaiPCD20}, and End-to-End ABSA (E2E-ABSA)~\cite{DBLP:conf/acl/HeLND19,DBLP:conf/emnlp/LiLWBZY19}, aspect-sentiment triplet extraction (ASTE)~\cite{DBLP:conf/aaai/ChenWLW21,DBLP:conf/emnlp/MukherjeeNBB021}, and Target Aspect Sentiment Detection (TASD)~\cite{DBLP:conf/aaai/WanYDLQP20}, have been proposed to determine one or several aspects of the sentiment.

Among all ABSA methods, aspect sentiment quad prediction (ASQP)~\cite{DBLP:conf/emnlp/ZhangD0YBL21} or aspect-category-opinion-sentiment (ACOS) quadruple extraction~\cite{DBLP:conf/acl/CaiXY20} is more critical because it needs to extract all aspect-category-opinion-sentiment quadruples present in a given sentence.  For example, given a comment for a car, ``The interior design is very delicate,'' the $(c, a, o, s)$-quadruple can be extracted to  (``Interior\#Interior Overall'', ``interior design'', ``very delicate'', POS), where POS is the positive sentiment polarity.  The task comprises four tasks, out of which two involve extracting aspect and opinion terms from the given sentence, while the other two involve classifying aspect category and opinion sentiment.

{
In this paper, we focus on ASQP because it offers a more nuanced understanding of the sentiment toward specific aspects or features of a product or services that can be highly beneficial in various applications~\cite{DBLP:journals/corr/abs-2203-01054}.  This task involves two main steps: extracting aspect and opinion terms and classifying them to the corresponding aspect categories and sentiment polarity.  Due to its practical significance, many researchers have devoted themselves to tackling it.  Typical methods can be divided into pipeline-based methods~\cite{DBLP:conf/acl/CaiXY20} and generation-based methods~\cite{DBLP:conf/ijcai/BaoWJXL22,DBLP:conf/coling/GaoFLLLLBY22,DBLP:conf/acl/MaoSYZC22,DBLP:conf/emnlp/ZhangD0YBL21}. 

However, technical exploration of ASQP is challenged by the small size and low quadruple density of current datasets.  To expand capacity, we curate and construct two datasets, \textbf{Car-ASQP} and \textbf{Digital-ASQP}. These datasets contain over 10,000 samples, which is 2.6 to 4.8 times larger than existing ASQP datasets. In addition, the released datasets have a higher density of quadruples, words per aspect, and words per opinion (2.2 to 3.9 times higher) and contain more words per sentence (9.7 to 11.0 times more) than the existing datasets. The released datasets are in Chinese and have complete quadruple annotations, enabling research on ASQP in multilingual contexts and the possibility of exploring other ABSA tasks, such as ACD, AOPE, ASTE, TASD, and more.

The Generative Pre-trained Transformer (GPT)~\cite{DBLP:conf/nips/BrownMRSKDNSSAA20} series of models have garnered interest for their superior performance in processing natural language processing tasks.  In this paper, we are the first to evaluate ASQP performance on GPT series models and observe poor performance in the zero-shot learning setting.  Further investigation under a redesigned aspect-opinion-sentiment triple prediction reveals that F1 scores increase by 2.1 to 8.6 times when relaxing the measurement of the correctness of aspect and opinion terms. The results identify an issue distinguishing between aspect and opinion terms in GPT series models.  The comparison with strong baselines underscores the need for further investigation into techniques for ASQP and potential improvements in GPT series models.



}

{

}

\begin{CJK*}{UTF8}{gbsn}
\begin{figure}[htpb]
\scriptsize
\centering
\begin{tabular}{cc}
\hline
Opinionated sentences & Quadruples \\\hline
\begin{minipage}{3.4cm}
\begin{tabular}{p{3.4cm}}
\textcolor{aspect}{内饰}\textcolor{opinion}{差}、\textcolor{aspect}{屏幕}不大。\textcolor{opinion}{不要}\textcolor{aspect}{购买}
\\  \textcolor{aspect}{Interior} is \textcolor{opinion}{poor}, \textcolor{aspect}{screen} is \textcolor{opinion}{not large}. \textcolor{opinion}{Don't }\textcolor{aspect}{purchase} it
\end{tabular}
 \end{minipage} 
 & 
 \begin{minipage}{8.5cm}
\begin{tabular}{p{8.5cm}}
(\textcolor{category}{内饰\#内饰整体}, \textcolor{aspect}{内饰}, \textcolor{opinion}{差}, \textcolor{sentiment}{NEG})\\
(\textcolor{category}{内饰\#车载屏幕}, \textcolor{aspect}{屏幕}, \textcolor{opinion}{不大}, \textcolor{sentiment}{NEG})\\
(\textcolor{category}{买家态度\#购买意愿}, \textcolor{aspect}{购买}, \textcolor{opinion}{不要}, \textcolor{sentiment}{NEG})\\
(\textcolor{category}{Interior\#Interior Overall}, \textcolor{aspect}{Interior}, \textcolor{opinion}{poor}, \textcolor{sentiment}{NEG})\\
(\textcolor{category}{Interior\#Car Screen}, \textcolor{aspect}{screen}, \textcolor{opinion}{not large}, \textcolor{sentiment}{NEG})\\
(\textcolor{category}{Buyer\'s Attitude\#Willingness To Buy}, \textcolor{aspect}{Don't purchase}, \textcolor{opinion}{NULL}, \textcolor{sentiment}{NEG})
\end{tabular}
 \end{minipage}
\\ \hline
\begin{minipage}{3.4cm}
\begin{tabular}{p{3.4cm}}
\textcolor{aspect}{屏幕色彩分辨度}\textcolor{opinion}{高}。 \textcolor{aspect}{功能}\textcolor{opinion}{很齐全}, 最主要\textcolor{aspect}{便宜}\\
\textcolor{aspect}{The screen} \textcolor{opinion}{high} \textcolor{aspect}{color resolution}.  It has \textcolor{opinion}{very complete} \textcolor{aspect}{functions}, and most importantly, {it} is \textcolor{aspect}{cheap}
 \\\\
\end{tabular}
 \end{minipage}
 & 
\begin{minipage}{8.5cm}
\begin{tabular}{p{8.5cm}}
(\textcolor{category}{屏幕\#屏幕分辨率}, \textcolor{aspect}{分辨度}, \textcolor{opinion}{高}, \textcolor{sentiment}{POS})\\
(\textcolor{category}{屏幕\#屏幕色彩}, \textcolor{aspect}{屏幕色彩}, \textcolor{opinion}{高}, \textcolor{sentiment}{POS})\\
(\textcolor{category}{功能/应用\#功能/应用}, \textcolor{aspect}{功能}, \textcolor{opinion}{很齐全}, \textcolor{sentiment}{POS})\\
(\textcolor{category}{价值体验\#价格}, \textcolor{aspect}{便宜}, \textcolor{opinion}{NULL}, \textcolor{sentiment}{POS})\\
(\textcolor{category}{Screen\#Screen Resolution}, \textcolor{aspect}{Resolution}, \textcolor{opinion}{high}, \textcolor{sentiment}{POS})\\
(\textcolor{category}{Screen\#Screen Color}, \textcolor{aspect}{Screen Color}, \textcolor{opinion}{high}, \textcolor{sentiment}{POS})\\
(\textcolor{category}{Function/Application\#Function/Application}, \textcolor{aspect}{functions}, \textcolor{opinion}{very complete}, \textcolor{sentiment}{POS})\\
(\textcolor{category}{Value Experience\#Price}, \textcolor{aspect}{cheap}, \textcolor{opinion}{NULL}, \textcolor{sentiment}{POS})
 \end{tabular}
 \end{minipage}
\\ \hline
\end{tabular}
\caption{Typical examples of the labeled quadruples.
\label{tab:ex_illustration}}
\end{figure}
\end{CJK*}
\section{Datasets}
We construct two novel Chinese datasets, Car-ASQP and Digital-ASQP, for ASQP.

\subsection{Source}
{
The {Car-ASQP} and {Digital-ASQP} datasets are collected from forums~\footnote{\url{https://club.autohome.com.cn/},\url{https://www.dgtle.com/}}, news~\footnote{\url{https://auto.huanqiu.com/},\url{https://www.zol.com.cn/}}, several e-commerce platforms~\footnote{\url{https://www.jd.com/},\url{https://www.taobao.com/}}, Little Red Booklittle~\footnote{\url{https://www.xiaohongshu.com/}} and  Weibo~\footnote{\url{https://weibo.com/}} in the years 2019-2022 under the categories of Car and Digital.  We select these two categories because they contain detailed customer comments.  We clean the data by the following steps: (1) filtering out the samples with lengths less than $10$ and greater than $512$; (2) filtering out the samples with valid Chinese characters less than 70\%; (3) filtering out ad texts by a classifier which is trained by marketing texts with 90\% classification accuracy. 
}

\subsection{Annotation}\label{sec:guideline}
The annotation procedure follows: Two annotators individually annotate the text by our developed labeled system.  The strict quadruple matching F1 score between two annotators is greater than 77\%, implying a substantial agreement between two annotators~\cite{DBLP:conf/coling/KimK18}.  When there is disagreement, the project leader will be asked to make the final decision.  

The annotation guideline is as follows: 
\begin{compactitem}
\item The \textbf{aspect category} is defined by a two-level label-category system, which our business experts determine for commercial value and more detailed information.  For example, the first-level category of ``Interior'' in Car-ASQP includes the second-level categories, such as ``Interior Overall'', ``Central Control Area'', ``Seat'', ``Air Conditioner'', etc.  We can then get the final categories by concatenating them via \#, e.g., ``Interior\#Seat''.  
\item The \textbf{aspect term} is usually a noun or a noun phrase, explicitly or implicitly linking to the opinion target in the text~\cite{DBLP:conf/acl/CaiXY20}.  For commercial consideration, we exclude sentences {\em without} aspects.  Moreover, to provide more fine-grained information, we include three additional rules: 
\begin{compactitem}
    \item The aspect term can be an adjective or verb when it reveals the sentiment category.  For example, as shown in Table~\ref{tab:ex_illustration}, ``cheap'' is labeled as an aspect term because it specifies the category of ``Value Experience\#Price''.
    \item Pronoun is not allowed to be an aspect term as it lacks sufficient context.    
    \item Top-priority in labeling fine-grained aspects.  For example, in ``Don't purchase it'', ``purchase'' relates more to a customer's purchasing while “product” is more related to the overall comment,.  We label ``purchase'' as the aspect.  
\end{compactitem}

\item An {\bf opinion term} is usually an adjective or a phrase with sentiment polarity.  Here, we include more labeling criteria:
\begin{compactitem}
    \item When there is a negative word, e.g., ``Don't'', ``NO'', ``cannot'', ``doesn't'', the negative word should be included in the opinion term.  For example, ``not large'' is labeled as the corresponding opinion terms in Table~\ref{tab:ex_illustration}.
    \item When there is an adverb or a preposition, e.g., ``very'', ``too'', ``so'', ``inside'', ``under'', or ``outside'', the corresponding adverb or preposition should be included in the opinion term.  For example, in Table~\ref{tab:ex_illustration}, ``very complete'' is labeled as an  opinion term.  Usually, in Restaurant-ACOS and Laptop-ACOS, ``very'' is not included in the opinion term.  This increases the difficulty of extracting opinion terms and demonstrates the significance of our released datasets to the ASBA community.
\end{compactitem}

\item The \textbf{sentiment polarity} belongs to one of the sentiment classes, \{POS, NEU, NEG\}, for the positive, neutral, and negative sentiment, respectively.  
\end{compactitem}

\subsection{Statistics and Analysis}
Table~\ref{tab:datasets} reports the statistics of two existing English ASQP datasets and our released Chinese ASQP datasets.  Our released datasets contain the following characteristics:
\begin{compactitem}
    \item {\bf larger size and rich aspect categories}: The released datasets consist of 10,000+ samples each, which is 2.6 to 4.8 times larger than the existing ASQP datasets.  Additionally, the number of categories in the released datasets is significantly greater than those in the Restaurant-ACOS and comparable to the Laptop-ACOS dataset, which has 121 categories, reduced to 60 by filtering categories with less than ten annotated quadruples.
    \item {\bf more words per sentence}: The fourth row of the table indicates that the released datasets contain 9.7 to 11.0 times more words per sentence than the existing ASQP datasets.     
    \item {\bf higher density}: Rows five through seven of the table show that the released datasets have 2.2 to 3.9 times higher density in terms of quadruples, words per aspect, and words per opinion compared to the existing ASQP datasets.
\end{compactitem}
Overall, our recently released Chinese ASQP datasets are more detailed and comprehensive than the existing ones, intending to promote technical advancements in this field.

\begin{table}[htp]
\scriptsize
\centering\caption{Data statistics.  \# denotes the number of corresponding elements.  s, w, $c_1$/$c_2$, q represent samples, words, first/second-level categories, and quadruples.  EA, EO, and IO denote explicit aspect, explicit opinion, and implicit opinion, respectively.  The average value is denoted via overline, e.g., $\overline{\#\mbox{w}/\mbox{s}}$ is the average of words per sentence. } 
\label{tab:datasets}
\begin{tabular}{
l@{~~~~}c@{~~~~}c@{~~~~}c@{~~~~}c}
\hline
& Restaurant-ACOS & Laptop-ACOS & Car-ASQP & Digital-ASQP \\\hline
{\#s} & 2,286 & 4,076 & 10,894 & 10,580\\
\#$c_1$/\#$c_2$ & 13/- & 23/121 & 24/90 & 32/118 \\
{\#q} & 3,661 & 5,773 & 39,075 & 36,741 \\
$\overline{\#\mbox{w}/\mbox{s}}$ & 15.11 & 15.73 & 166.88 & 152.26 \\
$\overline{\#\mbox{q}/\mbox{s}}$ & 1.60 & 1.42 & 3.59 & 3.47\\ 
$\overline{\#\mbox{w}/a}$ 
& 1.46 & 1.40 & 3.26 & 3.15 \\ 
$\overline{\#\mbox{w}/o}$
& 1.20 & 1.09 & 4.20 & 3.41 \\ 
EA\&EO & 350 & 342 & 30,591 & 34,055\\
EA\&IO & 350 & 1,241 & 8,484 & 2,686\\ 
IA\&EO & 530 & 912 & - & - \\ 
IA\&IO & 2,431 & 3,278 & - & - \\ 
{\#NEG} & 1,007 & 1,879 & 4,784 & 10,396\\ 
{\#NEU} & 151 & 316 & 18,284 & 3,710\  \\ 
{\#POS} &  2,503 & 3,578 & 16,007 & 22,635 \\
\hline
\end{tabular}
\end{table}


\begin{CJK*}{UTF8}{gbsn}
\begin{figure}[htpb]
\scriptsize
\centering
\begin{tabular}{ccc}
\hline
\begin{minipage}{7.50cm}
\begin{tabular}{p{7.50cm}}
  You are an analyst who needs to perform sentiment analysis on online comments about a product across four dimensions: viewpoint category, feature word, sentiment word, and sentiment polarity.\\\vspace{0.05cm}
  \textcolor{category}{{Viewpoint category}: \{category\_desc\}}\\
  \textcolor{aspect}{Feature word: a word that determines the evaluation dimension, such as ``quality" and ``price".  For example, if evaluating a product on Amazon, the feature words could be ``quality," ``price," or ``logistics." ``Quality," ``material," and ``workmanship" are feature words for the dimension of quality; ``discounted price," ``price," ``cost," and ``selling price" are feature words for the dimension of price; and ``delivery," ``packaging," and ``logistics" are feature words for the dimension of logistics.  If there is no feature word, fill in the sentiment word.}\\
  \textcolor{opinion}{Sentiment word: the sentiment of the evaluated feature word. If there is no sentiment word, fill in NULL.}\\
\textcolor{sentiment}{Sentiment polarity: POS, NEU, or NEG.}\\\vspace{0.05cm}
For example:\\
{\bf Input}:The cost-effectiveness is quite high, and the appearance is also good-looking. The key is that the family is satisfied with this car
 \\
{\bf Output}: [[Price, cost-effectiveness, quite high, POS], [Appearance, appearance, good-looking, POS], [Product Perception, satisfied, NULL, POS]]\\\vspace{0.05cm}

Input: \{input\_text\}\\ 
Output:
  \end{tabular}
\end{minipage}
& & 
\begin{minipage}{4.50cm}
\begin{tabular}{p{4.50cm}}
你是一个分析师, 现在要对网民针对产品的评论进行四个维度的情感分析: 观点类型, 特征词, 情感词, 情感极性\\\vspace{0.05cm}
  \textcolor{category}{观点类型: \{category\_desc\}}\\
  \textcolor{aspect}{特征词: 确定评价维度的词, 如“质量”、“价格”或“物流”。例如，如果在亚马逊上评估产品，特征词可能是“质量”、“价格”或“物流”。 “质量”、“材料”、“工艺”是质量维度的特征词；“折扣价”、“性价比”、“成本”、“售价”是价格维度的特征词。 价格维度；而“交付”、“包装”和“物流”是物流维度的特征词。 如果没有特征词，就填写情感词。}
  \\
  \textcolor{opinion}{情感词：被评价的特征词的情感。 如果没有情感词，则填NULL。}\\
  \textcolor{sentiment}{情感极性：正面、中性或负面。}\\\vspace{0.05cm}
    样例：\\
    {\bf 输入}：性价比比较高吧颜值也可以。关键是家人满意这个车。\\
{\bf 输出}：[[价格, 性价比, 比较高, 正面], [外观, 颜值, 也可以, 正面], [产品感知, 满意, NULL, 正面]]
    
\\\vspace{0.05cm}

{输入}：\{input\_text\}\\
输出：
 \end{tabular}
 \end{minipage}
\\ \hline
\end{tabular}
\caption{Prompts for GPT series models in both English and Chinese.  \{category\_desc\} describes the aspect category, and \{input\_text\} is the input text; see more descriptions in the main text. 
\label{fig:gpt_prompt}}
\end{figure}
\end{CJK*}

\section{Experiments}

We benchmark the ASQP task with two GPT series models and three strong baselines: 
\begin{compactitem}
    \item \textbf{Extract-Classify-ACOS}~\cite{DBLP:conf/acl/CaiXY20}: the best pipeline-based method for ASQP by first performing aspect-opinion co-extraction and following category-sentiment prediction;
    \item {\bf Paraphrase} generation for ASQP~\cite{DBLP:conf/emnlp/ZhangD0YBL21}: the first and a strong generation-based method for ASQP;
    \item \textbf{GEN-SCL-NAT}~\cite{DBLP:conf/emnlp/Peper022}: exploiting supervised contrastive learning and a new structured generation format to improve the naturalness of the output sequences for ASQP; 
\end{compactitem}

\paragraph{Settings and Evaluation Metrics} 
{
We conduct the experiments on four datasets in Table~\ref{tab:datasets}.  For Restaurant-ACOS and Laptop-ACOS, we apply the original splitting on the training, validation, and test sets~\cite{DBLP:conf/acl/CaiXY20}.  For Car-ASQP and Digital-ASQP, we split the corresponding datasets via 7:1.5:1.5.  We employ F1 scores as the primary evaluation metric and report the corresponding \textbf{P}recision and \textbf{R}ecall scores.  To be considered correct, a sentiment quad prediction must match the gold labels exactly in all predicted elements.  
}

\paragraph{Implementation Details}  
We adopt PyTorch 1.13.1 to run the baselines on a high-performance workstation with an Intel Xeon CPU, 256GB memory, a single A5000 GPU, and Ubuntu 20.04.  In Extract-Classify-ACOS, we adopt bert-base-uncased~\cite{DBLP:conf/naacl/DevlinCLT19} for English datasets and macbert-base~\cite{DBLP:conf/emnlp/CuiC000H20} for Chinese datasets  task.  For Paraphrase and GEN-SCL-NAT, we apply T5-base~\cite{DBLP:journals/jmlr/RaffelSRLNMZLL20} for English datasets and Mengzi-t5-base~\cite{DBLP:journals/corr/abs-2110-06696} for Chinese datasets. The learning rate and batch size were set to [2e-5, 32] and [8e-5, 4], respectively to attain good performance. The maximum token length was set to 128 for English and 512 for Chinese datasets. We run each method for 20 epochs and evaluate them on the test set when the models performed best on the validation set.

{
Two GPT series models, text-davinci-003 and gpt-3.5-turbo, are evaluated on the test datasets, with a typical prompt shown in Fig.~\ref{fig:gpt_prompt} for English and Chinese, respectively.  We set the temperature to 0, which means GPT series models will take fewer risks when making the prediction. 
Due to token length limitations, only the first-level aspect category is evaluated.  As shown in Fig.~\ref{fig:gpt_prompt}, the original second-level aspect categories, namely Price\#Cost Performance, Appearance\#Overall, and Product Perception\#Overall are replaced by their respective first-level aspect categories, namely Price, Appearance, and Product Perception.  We create \{category\_desc\} by concatenating all aspect categories separated by a comma and letting the GPT models identify aspect categories automatically.  For example, suppose Price, Appearance, and Product Perception are all aspect categories, the resulting \{category\_desc\} would be ``Price,Appearance,Product Perception". 

}

\subsection{Main Results}

\begin{table}[htpb]
\scriptsize
\centering\caption{Overall performance on all four datasets.  Scores are averaged over five runs with different seeds for the three baselines. }
\label{tab:main_result}
\begin{tabular}{p{2cm}|ccc|ccc|ccc|ccc}
\hline
\multirow{2}{*}{\textbf{Method}}
& \multicolumn{3}{c|}{\textbf{Restaurant-ACOS}} & \multicolumn{3}{c|}{\textbf{Laptop-ACOS}} & \multicolumn{3}{c|}{\textbf{Car-ASQP}} & \multicolumn{3}{c}{\textbf{Digital-ASQP}} \\ \cline{2-13} 
& {{P}} & {{R}} & {{F1}} & {{P}} & {{R}} & {{F1}} & {{P}} & {{R}} & {{F1}} & {{P}} & {{R}} & {{F1}} \\ \hline
Extract-Classify & 38.54 & 52.96 & 44.61 & {\textbf{45.56}} & 29.48 & 35.80 & 10.82 & {\textbf{38.79}} & 16.92 & 14.65 & 32.18 & 20.14 \\ 
Paraphrase & 58.98 & 59.11 & 59.04 & 41.77 & 45.05 & 43.34 & 33.27 & 28.76 & 30.85 & 40.93 & 36.02 & 38.32 \\
GEN-SCL-NAT & {\textbf{60.23}} & {\textbf{59.34}} & {\textbf{59.78}} & 42.49 & {\textbf{45.98}} & {\textbf{44.16}} & {\textbf{35.59}} & 30.64 & {\textbf{32.93}} & {\textbf{42.00}} & {\textbf{41.46}} & {\textbf{41.73}} \\
\hline
text-davinci-003 & 15.95 & 20.02 & 17.76 & 6.13 & 8.33& 7.06  & 4.05 & 3.96 & 4.01 & 4.06 & 4.04 & 4.05\\ 
gpt-3.5-turbo & 20.24 & 26.60 & 22.99 & 6.58 & 9.31 & 7.71  & 3.95 & 3.38 & 3.64 & 5.28 & 4.17 & 4.66\\\hline
\end{tabular}
\end{table}

{
Table~\ref{tab:main_result} reports the overall performance on all four datasets.  The results indicate that: (1) GEN-SCL-NAT attains the best performance among all compared methods, while all three baselines perform lower on the Chinese datasets than the English ones. We conjecture this is due to our released datasets' more affluent categories and higher density.  Further investigations and technical development can be explored in the future. (2) GPT series models perform poorly on all datasets, notably the released ones. However, this comparison is relatively unfair as GPT series models perform zero-shot learning without any labeled samples, and the categories are not explicitly provided for the models.  This is the same as those observed in general information extraction tasks~\cite{DBLP:journals/corr/abs-2304-11633,DBLP:journals/corr/abs-2304-08085}. In Sec.~\ref{sec:AOS_Rs} and Sec.~\ref{sec:error}, we explore more underlying reasons behind the models.
}

\begin{table}[htpb]
\scriptsize
\centering\caption{Redesigned aspect-opinion-sentiment performance (F1) to investigate text-davinci-003 and gpt-3.5-turbo. \label{tab:AOS_result}}
\begin{tabular}{l|cc|cc|cc|cc}
\hline
{\multirow{2}{*}{\textbf{Method}}} & \multicolumn{2}{c|}{\textbf{Restaurant-ACOS}} & \multicolumn{2}{c|}{\textbf{Laptop-ACOS}} & \multicolumn{2}{c|}{\textbf{Car-ASQP}} & \multicolumn{2}{c}{\textbf{Digital-ASQP}} \\ \cline{2-9} 
 & {{$d=0$}} & {{$d \leq 5$}}  & {{$d=0$}} & {{$d \leq 5$}} & {{$d=0$}} & {{$d \leq 2$}} & {{$d=0$}} & {{$d \leq 2$}} \\ \hline
Extract-Classify & 50.02 & 66.70 & 48.62 & 53.81 & 18.85 & 52.04 & 22.08 & 80.70 \\
Paraphrase & 64.38 & 71.59 & 68.20 & 75.02 & 34.76 & 46.63 & 43.94 & 57.28 \\
GEN-SCL-NAT & {\textbf{65.32}} & {\textbf{72.39}} & {\textbf{69.23}} & {\textbf{75.32}} & {\textbf{35.16}} & {\textbf{46.93}} & {\textbf{46.48}} & {\textbf{61.52}} \\\hline
text-davinci-003 & 24.88 & 47.51 & 16.64 & 28.08 & 9.73 & 26.65 & 16.83 & 34.99 \\
gpt-3.5-turbo & 32.71 & 47.21 & 16.15 & 28.81 & 8.34 & 21.86 & 11.71 & 27.58 \\\hline
\end{tabular}
\end{table}
\subsection{Redesigned Aspect-Opinion-Sentiment Evaluation}\label{sec:AOS_Rs}
We redesign an experiment to evaluate the aspect-opinion-sentiment (AOS) performance to understand the underlying reasons behind the poor performance of text-davinci-003 and gpt-3.5-turbo.  Due to the large size of the aspect categories set and the unfairness of GPT series models, we exclude its evaluation from our analysis.  Additionally, we observe that the aspect and opinion may be correctly extracted but not well-separated due to prompt misunderstandings.  To address this, we deem an aspect-opinion-sentiment triple prediction is correct when the sentiment polarity is correct and the following edit distance $d$ is 0 or less than a threshold:
\begin{equation}\label{eq:ed_ao}
d = \min\quad \{d(p_{ao}, t_{ao}), d(p_{oa}, t_{oa})\},
\end{equation}
where $p$ and $t$ represent the predicted and ground truth values, respectively.  $ao$ and $oa$ denote the concatenations of $a$ and $o$. 

\begin{CJK*}{UTF8}{gbsn}
{
The results presented in Table~\ref{tab:AOS_result} highlight two key findings.  Firstly, GEN-SCL-NAT consistently performs best among all compared methods and outperforms its corresponding ASQP task, especially when the number of aspect categories is larger. This suggests that classifying aspect categories remains a challenging task for ASQP. Secondly, for the GPT series models, the F1 scores increase as $d$ approaches 0. Specifically, when $d \leq 5$ for English datasets and $d \leq 2$ for Chinese datasets, the F1 scores increase by 2.1 to 8.6 times, respectively.  These results indicate that a common issue with GPT series models is the unclear boundary between aspect and opinion terms. By relaxing the metric used to measure the correctness of these terms, the performance of GPT series models can be improved.  More exploration can be conducted in the future work.


} 
\end{CJK*}

\subsection{Effect of Handling Implicit Opinions}\label{sec:IO}
For three reasons, we report the implicit opinion breakdown performance only in the baselines: (1) Car-ASQP and Digital-ASQP lack implicit aspects, (2) GPT series models perform poorly and are unfair for comparison, and (3) space limitations exist.  The results in Table~\ref{tab:implicit} reveal three key findings: (1) Generation-based methods are more effective in handling implicit opinions than pipeline-based methods; (2) GEN-SCL-NAT consistently outperforms Paraphrase across all datasets; (3) The performance on EA\&IO is worse than EA\&EO, indicating the challenge of handling implicit opinions.

\begin{table}[htpb]
\scriptsize
\centering\caption{Breakdown performance for handling implicit opinions on baselines.}  
\label{tab:implicit}
\begin{tabular}{l|cc|cc|cc|cc}
\hline
{\multirow{2}{*}{\textbf{Method}}} & \multicolumn{2}{c|}{\textbf{Restaurant-ACOS}} & \multicolumn{2}{c|}{\textbf{Laptop-ACOS}} & \multicolumn{2}{c|}{\textbf{Car-ASQP}} & \multicolumn{2}{c}{\textbf{Digital-ASQP}} \\ \cline{2-9} 
 & {{EA\&EO}} & {{EA\&IO}}  & {{EA\&EO}} & {{EA\&IO}} & {{EA\&EO}} & {{EA\&IO}} & {{EA\&EO}} & {{EA\&IO}} \\ \hline
Extract-Classify & 45.0 & 23.9 & 35.4 & 16.8 & 16.60 & 16.31  & 20.25 & 17.31 \\
Paraphrase & 62.14 & 41.8 & 43.6 & 32.5  & 30.76 & 30.21 & 38.70 & 32.13 \\
GEN-SCL-NAT & {\textbf{64.9}} & {\textbf{43.2}} & {\textbf{44.2}} & {\textbf{33.6}} & {\textbf{33.01}} & {\textbf{32.63}} & {\textbf{42.01}} & {\textbf{35.33}} \\\hline
\end{tabular}
\end{table}

\begin{figure}[htp]
\subfigure[text-davinci-003]{\includegraphics[scale=0.05]{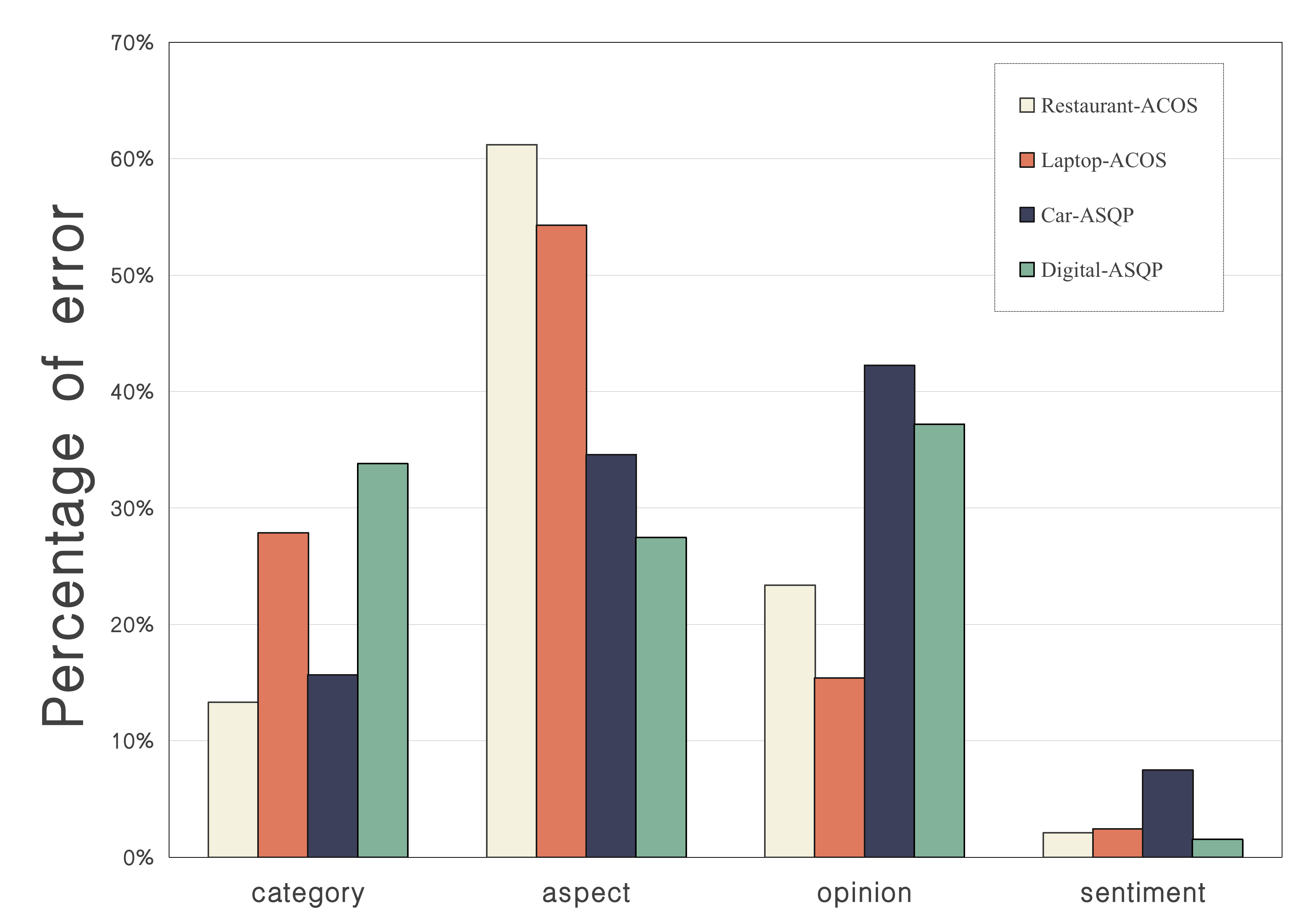}} 
\subfigure[gpt-3.5-turbo]{\includegraphics[scale=0.05]{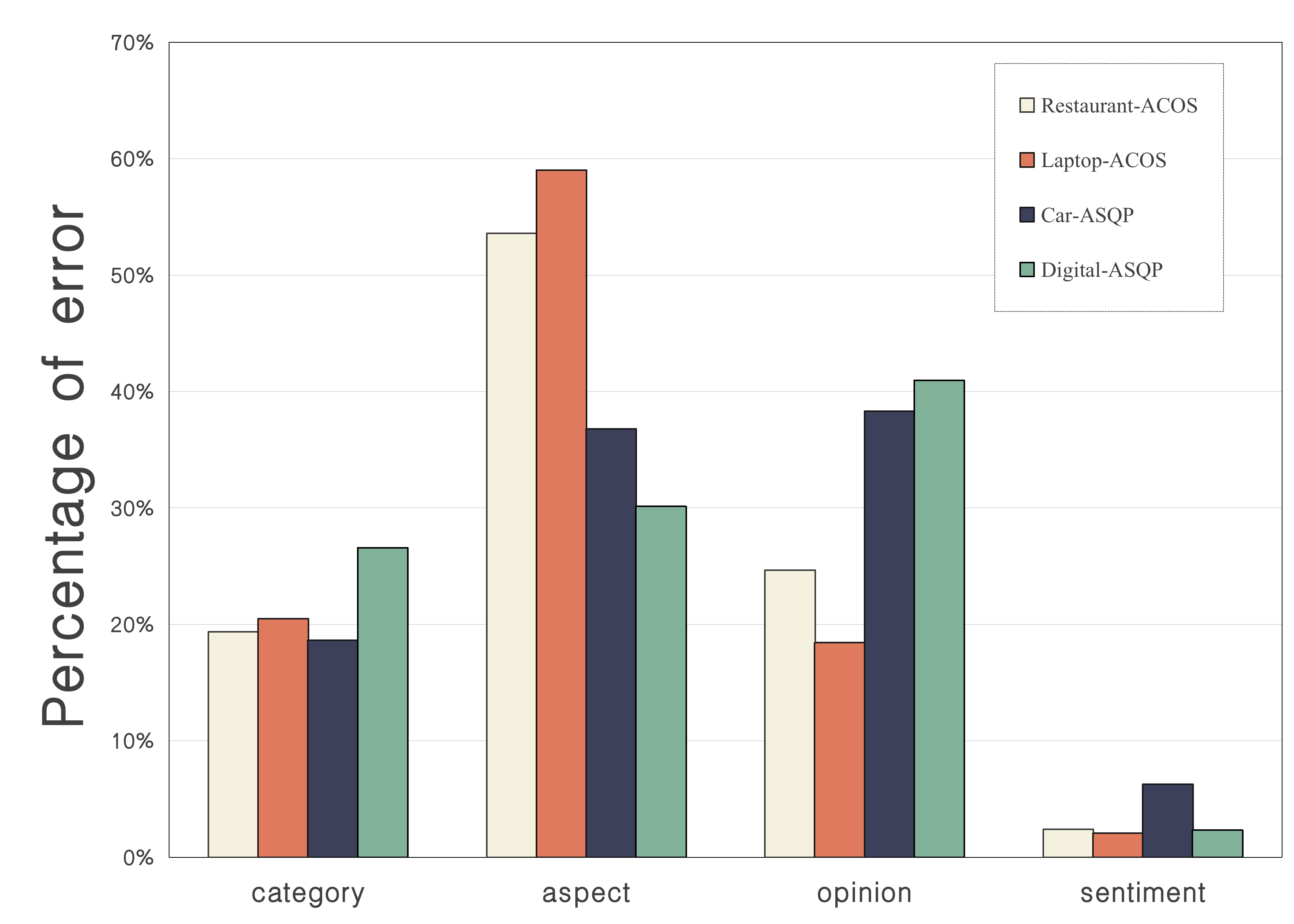}} 
\caption{Breakdown error analysis of GPT series models.\label{fig:error_analysis} }
\end{figure}
\subsection{Error Analysis and Case Study for GPT Series Models}\label{sec:error}
\begin{CJK*}{UTF8}{gbsn}
{


Here, we conducted error analysis and case study to better understand GPT series models' characteristics, particularly when they fail.  We examine the incorrect quad predictions on all datasets and report the percentage of errors for better illustration in Fig.~\ref{fig:error_analysis}. The results indicate that text-davinci-003 and gpt-3.5-turbo models struggle to extract aspect terms in English datasets due to the implicit aspects issue. Their performance declines in extracting opinion terms in the released datasets.  However, in classification tasks, GPT models achieve lower error rates, less than 7.5\%, in sentiment identification. The above findings suggest the need for further exploration in prompt design to improve the classification of aspect categories and extraction of aspect and opinion terms.

{ Fig.~\ref{fig:case_study} provides some typical failures of gpt-3.5-turbo in Digital-ASQP.  Regarding category, gpt-3.5-turbo is often confused by semantically similar categories, such as ``Human-machine Interaction" and ``Function/Applications." This suggests the need for more category definitions in the prompt.  We also identified boundary errors in extracting aspects and opinions, as demonstrated by the opinion error where the model missed the phrase ``a bit".  Regarding sentiment errors, gpt-3.5-turbo tends to confuse NEU and NEG, which may be attributed to the sentiment imbalance in Digital-ASQP.  Specifically, we observe that NEU data accounts for around 10\%.}



}
\end{CJK*}

\begin{CJK*}{UTF8}{gbsn}
\begin{figure}[htpb]
\scriptsize
\begin{tabular}{|@{~}l@{~}|l@{~}|}
\hline
\textbf{Type} & \textbf{Example} \\\hline
\multirow{6}{*}{Category} & Input: 十次有九次不答应！语音识别太菜	\\
& Gold: (\underline{\blue{人机交互}}, 语音识别, 太菜, 负面) \\
& Pred: (\underline{\blue{功能/应用}}, 语音识别, 太菜, 负面) \\
& Input: Nine times out of ten, no! Voice recognition is too terrible	\\
& Gold: (\underline{\blue{Human-computer Interaction}}, Voice recognition, too terrible, NEG) \\
& Pred: (\underline{\blue{Functions/Applications}}, Voice recognition, too terrible, NEG) \\
\hline
\multirow{6}{*}{Aspect} 	& Input: 为什么我已经打开，并且播放了音乐锁屏之后还是没有特效出来？？		\\
& Gold: (功能/应用, \underline{\blue{锁屏}}, 没有特效出来, 负面) \\
& Pred: (功能/应用, \underline{\blue{音乐锁屏}}, 没有特效出来, 负面) \\
& Input: Why are there no effects after I have opened and played the music lock screen?		\\
& Gold: (Functions/Applications, \underline{\blue{lock screen}}, no effects, NEG) \\
& Pred: (Functions/Applications, \underline{\blue{the music lock screen}}, no effects, NEG) \\
\hline
\multirow{6}{*}{Opinion} & Input: 不好的地方是前置摄像头有点卡顿 \\
& Gold: (摄影, 前置摄像头, \underline{\blue{有点卡顿}}, 负面) \\
& Pred: (摄影, 前置摄像头, \underline{\blue{卡顿}}, 负面) \\
& Input: The bad part is that the front camera is a bit stuck \\
& Gold: (Photography, the front camera, \underline{\blue{a bit stuck}}, NEG) \\
& Pred: (Photography, the front camera, \underline{\blue{stuck}}, NEG) \\
\hline
\multirow{6}{*}{Sentiment} & Input: 前置摄像头左边自己会动	\\
& Gold: (摄影, 前置摄像头, 自己动, \underline{\blue{负面}}) \\
& Pred: (摄影, 前置摄像头, 自己动, \underline{\blue{中性}}) \\
& Input: The left side of the front camera will move by itself	\\
& Gold: (Photography, the front camera, move by itself, \underline{\blue{NEG}}) \\
& Pred: (Photography, the front camera, move by itself, \underline{\blue{NEU}}) \\
\hline 
\end{tabular}         
\caption{Case study of gpt-3.5-turbo in Digital-ASQP.} 
\label{fig:case_study} 
\end{figure}
\end{CJK*}

\section{Related Work}


The {\bf ABSA benchmark datasets} are primarily sourced from the SemEval'14-16 shared challenges~\cite{DBLP:conf/semeval/PontikiGPAMAAZQ16,DBLP:conf/semeval/PontikiGPMA15,DBLP:conf/semeval/PontikiGPPAM14}, with the original task being the identification of opinions related to specific entities and their aspects. To explore additional tasks, such as AOPE, E2E-ABSA, ASTE, TASD, and ASQP, researchers have re-annotated existing datasets and created new ones.  However, re-annotated datasets have limitations, including small data sizes, low quadruple density, and only in English. These shortcomings have prompted us to crawl and curate larger datasets with higher quadruple density.

The {\bf GPT}~\cite{DBLP:conf/nips/BrownMRSKDNSSAA20} series models have recently gained significant attention due to their exceptional performance in unifying all NLU tasks into generative tasks.  Various empirical evaluations have been conducted to examine the GPT models' performance, e.g., in coreference resolution~\cite{DBLP:conf/acl-insights/YangPMT22} and typical NLU tasks~\cite{DBLP:journals/corr/abs-2302-06476,DBLP:journals/corr/abs-2303-10420,DBLP:conf/blackboxnlp/ZhangWCZ22}. However, no testing has yet been performed on ASQP.

{
}

\section{Conclusions}
We have released two new Chinese datasets for ASQP and are the first to evaluate ASQP on GPT series models.  The released datasets are notable for their rich aspect categories, annotated with over 10,000 samples, offering around ten times more words per sentence and two to four times higher density than existing ASQP datasets.  This richness allows researchers to explore other subtasks in ABSA further.  Our experimental results underscore the need for further technological exploration to address ASQP and improve the performance of GPT models, for instance, through careful prompt design.

\bibliography{ref}

\begin{thebibliography}{10}
\providecommand{\url}[1]{\texttt{#1}}
\providecommand{\urlprefix}{URL }
\providecommand{\doi}[1]{https://doi.org/#1}

\bibitem{DBLP:conf/ijcai/BaoWJXL22}
Bao, X., Wang, Z., Jiang, X., Xiao, R., Li, S.: Aspect-based sentiment analysis
  with opinion tree generation. In: Raedt, L.D. (ed.) {IJCAI}. pp. 4044--4050
  (2022)

\bibitem{DBLP:conf/nips/BrownMRSKDNSSAA20}
Brown, T.B., Mann, B., Ryder, N., Subbiah, M., Kaplan, J., Dhariwal, P.,
  Neelakantan, A., Shyam, P., Sastry, G., Askell, A., Agarwal, S.,
  Herbert{-}Voss, A., Krueger, G., Henighan, T., Child, R., Ramesh, A.,
  Ziegler, D.M., Wu, J., Winter, C., Hesse, C., Chen, M., Sigler, E., Litwin,
  M., Gray, S., Chess, B., Clark, J., Berner, C., McCandlish, S., Radford, A.,
  Sutskever, I., Amodei, D.: Language models are few-shot learners. In: NeurIPS
  (2020)

\bibitem{DBLP:conf/coling/CaiTZYX20}
Cai, H., Tu, Y., Zhou, X., Yu, J., Xia, R.: Aspect-category based sentiment
  analysis with hierarchical graph convolutional network. In: {COLING}. pp.
  833--843 (2020)

\bibitem{DBLP:conf/acl/CaiXY20}
Cai, H., Xia, R., Yu, J.: Aspect-category-opinion-sentiment quadruple
  extraction with implicit aspects and opinions. In: {ACL/IJCNLP}. pp. 340--350
  (2021)

\bibitem{DBLP:conf/aaai/ChenWLW21}
Chen, S., Wang, Y., Liu, J., Wang, Y.: Bidirectional machine reading
  comprehension for aspect sentiment triplet extraction. In: {AAAI}. pp.
  12666--12674. {AAAI} Press (2021)

\bibitem{DBLP:conf/emnlp/CuiC000H20}
Cui, Y., Che, W., Liu, T., Qin, B., Wang, S., Hu, G.: Revisiting pre-trained
  models for chinese natural language processing. In: Findings of the
  Association for Computational Linguistics: {EMNLP}. Findings of {ACL}, vol.
  {EMNLP} 2020, pp. 657--668. Association for Computational Linguistics (2020)

\bibitem{DBLP:conf/emnlp/DaiPCD20}
Dai, Z., Peng, C., Chen, H., Ding, Y.: A multi-task incremental learning
  framework with category name embedding for aspect-category sentiment
  analysis. In: {EMNLP}. pp. 6955--6965 (2020)

\bibitem{DBLP:conf/naacl/DevlinCLT19}
Devlin, J., Chang, M., Lee, K., Toutanova, K.: {BERT:} pre-training of deep
  bidirectional transformers for language understanding. In: {NAACL-HLT}. pp.
  4171--4186 (2019)

\bibitem{DBLP:conf/coling/GaoFLLLLBY22}
Gao, T., Fang, J., Liu, H., Liu, Z., Liu, C., Liu, P., Bao, Y., Yan, W.:
  {LEGO-ABSA:} {A} prompt-based task assemblable unified generative framework
  for multi-task aspect-based sentiment analysis. In: {COLING}. pp. 7002--7012
  (2022)

\bibitem{DBLP:conf/acl/HeLND17}
He, R., Lee, W.S., Ng, H.T., Dahlmeier, D.: An unsupervised neural attention
  model for aspect extraction. In: {ACL}. pp. 388--397 (2017)

\bibitem{DBLP:conf/acl/HeLND19}
He, R., Lee, W.S., Ng, H.T., Dahlmeier, D.: An interactive multi-task learning
  network for end-to-end aspect-based sentiment analysis. In: {ACL}. pp.
  504--515. Association for Computational Linguistics (2019)

\bibitem{DBLP:conf/coling/KimK18}
Kim, E., Klinger, R.: Who feels what and why? annotation of a literature corpus
  with semantic roles of emotions. In: {COLING}. pp. 1345--1359 (2018)

\bibitem{DBLP:journals/corr/abs-2304-11633}
Li, B., Fang, G., Yang, Y., Wang, Q., Ye, W., Zhao, W., Zhang, S.: Evaluating
  chatgpt's information extraction capabilities: An assessment of performance,
  explainability, calibration, and faithfulness. CoRR  \textbf{abs/2304.11633}
  (2023)

\bibitem{DBLP:conf/emnlp/LiL17}
Li, X., Lam, W.: Deep multi-task learning for aspect term extraction with
  memory interaction. In: {EMNLP}. pp. 2886--2892 (2017)

\bibitem{DBLP:conf/emnlp/LiLWBZY19}
Li, Z., Li, X., Wei, Y., Bing, L., Zhang, Y., Yang, Q.: Transferable end-to-end
  aspect-based sentiment analysis with selective adversarial learning. In:
  {EMNLP-IJCNLP}. pp. 4589--4599 (2019)

\bibitem{DBLP:series/synthesis/2012Liu}
Liu, B.: Sentiment Analysis and Opinion Mining. Synthesis Lectures on Human
  Language Technologies, Morgan {\&} Claypool Publishers (2012)

\bibitem{DBLP:conf/emnlp/0030TCL021}
Liu, J., Teng, Z., Cui, L., Liu, H., Zhang, Y.: Solving aspect category
  sentiment analysis as a text generation task. In: {EMNLP}. pp. 4406--4416
  (2021)

\bibitem{DBLP:conf/acl/MaoSYZC22}
Mao, Y., Shen, Y., Yang, J., Zhu, X., Cai, L.: Seq2path: Generating sentiment
  tuples as paths of a tree. In: Findings of {ACL}. pp. 2215--2225 (2022)

\bibitem{DBLP:conf/emnlp/MukherjeeNBB021}
Mukherjee, R., Nayak, T., Butala, Y., Bhattacharya, S., Goyal, P.: {PASTE:} {A}
  tagging-free decoding framework using pointer networks for aspect sentiment
  triplet extraction. In: {EMNLP}. pp. 9279--9291 (2021)

\bibitem{DBLP:conf/emnlp/Peper022}
Peper, J., Wang, L.: Generative aspect-based sentiment analysis with
  contrastive learning and expressive structure. In: Findings of the
  Association for Computational Linguistics: {EMNLP}. pp. 6089--6095 (2022)

\bibitem{DBLP:conf/semeval/PontikiGPAMAAZQ16}
Pontiki, M., Galanis, D., Papageorgiou, H., Androutsopoulos, I., Manandhar, S.,
  Al{-}Smadi, M., Al{-}Ayyoub, M., Zhao, Y., Qin, B., Clercq, O.D., Hoste, V.,
  Apidianaki, M., Tannier, X., Loukachevitch, N.V., Kotelnikov, E.V., Bel, N.,
  Zafra, S.M.J., Eryigit, G.: Semeval-2016 task 5: Aspect based sentiment
  analysis. In: SemEval@NAACL-HLT. pp. 19--30. The Association for Computer
  Linguistics (2016)

\bibitem{DBLP:conf/semeval/PontikiGPMA15}
Pontiki, M., Galanis, D., Papageorgiou, H., Manandhar, S., Androutsopoulos, I.:
  Semeval-2015 task 12: Aspect based sentiment analysis. In: SemEval@NAACL-HLT.
  pp. 486--495 (2015)

\bibitem{DBLP:conf/semeval/PontikiGPPAM14}
Pontiki, M., Galanis, D., Pavlopoulos, J., Papageorgiou, H., Androutsopoulos,
  I., Manandhar, S.: Semeval-2014 task 4: Aspect based sentiment analysis. In:
  SemEval@COLING. pp. 27--35 (2014)

\bibitem{DBLP:journals/corr/abs-2302-06476}
Qin, C., Zhang, A., Zhang, Z., Chen, J., Yasunaga, M., Yang, D.: Is chatgpt a
  general-purpose natural language processing task solver? CoRR
  \textbf{abs/2302.06476} (2023)

\bibitem{DBLP:journals/jmlr/RaffelSRLNMZLL20}
Raffel, C., Shazeer, N., Roberts, A., Lee, K., Narang, S., Matena, M., Zhou,
  Y., Li, W., Liu, P.J.: Exploring the limits of transfer learning with a
  unified text-to-text transformer. J. Mach. Learn. Res.  \textbf{21},
  140:1--140:67 (2020)

\bibitem{DBLP:conf/aaai/WanYDLQP20}
Wan, H., Yang, Y., Du, J., Liu, Y., Qi, K., Pan, J.Z.: Target-aspect-sentiment
  joint detection for aspect-based sentiment analysis. In: {AAAI}. pp.
  9122--9129 (2020)

\bibitem{DBLP:journals/corr/abs-2304-08085}
Wang, X., Zhou, W., Zu, C., Xia, H., Chen, T., Zhang, Y., Zheng, R., Ye, J.,
  Zhang, Q., Gui, T., Kang, J., Yang, J., Li, S., Du, C.: Instructuie:
  Multi-task instruction tuning for unified information extraction. CoRR
  \textbf{abs/2304.08085} (2023)

\bibitem{DBLP:conf/emnlp/WuWP20}
Wu, M., Wang, W., Pan, S.J.: Deep weighted maxsat for aspect-based opinion
  extraction. In: {EMNLP}. pp. 5618--5628 (2020)

\bibitem{DBLP:conf/acl-insights/YangPMT22}
Yang, X., Peynetti, E., Meerman, V., Tanner, C.: What {GPT} knows about who is
  who. In: Insights@ACL. pp. 75--81. Association for Computational Linguistics
  (2022)

\bibitem{DBLP:journals/corr/abs-2303-10420}
Ye, J., Chen, X., Xu, N., Zu, C., Shao, Z., Liu, S., Cui, Y., Zhou, Z., Gong,
  C., Shen, Y., Zhou, J., Chen, S., Gui, T., Zhang, Q., Huang, X.: A
  comprehensive capability analysis of {GPT-3} and {GPT-3.5} series models.
  CoRR  \textbf{abs/2303.10420} (2023)

\bibitem{DBLP:journals/taslp/YuJX19}
Yu, J., Jiang, J., Xia, R.: Global inference for aspect and opinion terms
  co-extraction based on multi-task neural networks. {IEEE} {ACM} Trans. Audio
  Speech Lang. Process.  \textbf{27}(1),  168--177 (2019)

\bibitem{DBLP:conf/blackboxnlp/ZhangWCZ22}
Zhang, L., Wang, M., Chen, L., Zhang, W.: Probing gpt-3's linguistic knowledge
  on semantic tasks. In: BlackboxNLP@EMNLP. pp. 297--304. Association for
  Computational Linguistics (2022)

\bibitem{DBLP:conf/emnlp/ZhangD0YBL21}
Zhang, W., Deng, Y., Li, X., Yuan, Y., Bing, L., Lam, W.: Aspect sentiment quad
  prediction as paraphrase generation. In: {EMNLP}. pp. 9209--9219 (2021)

\bibitem{DBLP:journals/corr/abs-2203-01054}
Zhang, W., Li, X., Deng, Y., Bing, L., Lam, W.: A survey on aspect-based
  sentiment analysis: Tasks, methods, and challenges. CoRR
  \textbf{abs/2203.01054} (2022)

\bibitem{DBLP:journals/corr/abs-2110-06696}
Zhang, Z., Zhang, H., Chen, K., Guo, Y., Hua, J., Wang, Y., Zhou, M.: Mengzi:
  Towards lightweight yet ingenious pre-trained models for chinese. CoRR
  \textbf{abs/2110.06696} (2021)

\end{thebibliography}
\bibliographystyle{splncs04}
\end{document}